\documentclass[runningheads]{llncs}
\usepackage[T1]{fontenc}
\usepackage{adjustbox}
\usepackage{floatrow}
\usepackage{graphics}

\usepackage[font=small,labelfont=bf]{caption}

\usepackage{graphicx}
\usepackage{xcolor}
\usepackage{array}
\usepackage{amsmath}
\usepackage{amsfonts}
\usepackage{soul}
\usepackage{graphicx}
\usepackage{svg}
\begin{document}
\title{Longitudinal self-supervised learning using neural ordinary differential equation}

\titlerunning{Longitudinal self-supervised learning using NODE}

% \author{***************  \inst{1,2}
% % index{Li, Yihao}
% \and
% *************** \inst{1,3} % index{Conze,Pierre-Henri}
% \and
% ***************  \inst{1,2}%index{El Habib Daho, Mostafa}
% \and
% ***************  \inst{1,2}%index{Yihao, Li}
% \and
% *************** \inst{5}%index{El Habib Daho, Hugo}
% \and
% ***************  \inst{5} % index{Tadayoni, Ramin}
% \and
% ***************  \inst{5} % index{Massin, Pascal}
% \and
% ***************  \inst{1,2,4} % index{Cochener, Béatrice}
% \and
% *************** \inst{1,2,6} % index{Brahim, Ikram}
% \and 
% ***************  \inst{1} % index{Quellec, Gwenole}
% \and
% ***************  \inst{1,2} % index{Lamard, Mathieu}
% }

% \institute{
% ********************************   \and
% ******************************** \and
% ********************************
% \and
% ******************************** \and
% ******************************** \and
% ******************************** 
% }

% %\authorrunning{R. Zeghlache et al.}
% \authorrunning{*******************}
% % \institute{
% % LaTIM UMR 1101, Inserm, Brest, France \and
% % University of Western Brittany, Brest, France \and
% % IMT Atlantique, Brest, France
% % \and
% % Ophtalmology Department, CHRU Brest, Brest, France \and
% % Lariboisière Hospital, AP-HP, Paris, France
% % }
% %
% %

\author{Rachid Zeghlache\inst{1,2}
% index{Li, Yihao}
\and
Pierre-Henri Conze\inst{1,3} % index{Conze,Pierre-Henri}
\and
Mostafa El Habib Daho \inst{1,2}%index{El Habib Daho, Mostafa}
\and
Yihao Li \inst{1,2}%index{Yihao, Li}
\and
Hugo Le Boité \inst{5}%index{Hugo Le Boité}
\and
Ramin Tadayoni\inst{5} % index{Tadayoni, Ramin}
\and
Pascal Massin\inst{5} % index{Massin, Pascal}
\and
Béatrice Cochener\inst{1,2,4} % index{Cochener, Béatrice}
\and
Ikram Brahim \inst{1,2,6} % index{Brahim, Ikram}
\and 
Gwenolé Quellec\inst{1} % index{Quellec, Gwenole}
\and
Mathieu Lamard\inst{1,2} % index{Lamard, Mathieu}
}

\authorrunning{R. Zeghlache et al.}

\institute{
LaTIM UMR 1101, Inserm, Brest, France \and
University of Western Brittany, Brest, France \and
IMT Atlantique, Brest, France
\and
Ophtalmology Department, CHRU Brest, Brest, France \and
Lariboisière Hospital, AP-HP, Paris, France\and
LBAI UMR 1227, Inserm, Brest, France
}
\maketitle             % typeset the header of the contribution
\begin{abstract}

Longitudinal analysis in medical imaging is crucial to investigate the progressive changes in anatomical structures or disease progression over time. In recent years, a novel class of algorithms has emerged with the goal of learning disease progression in a self-supervised manner, using either pairs of consecutive images or time series of images. By capturing temporal patterns without external labels or supervision, longitudinal self-supervised learning (LSSL) has become a promising avenue. To better understand this core method, we explore in this paper the LSSL algorithm under different scenarios. The original LSSL is embedded in an auto-encoder (AE) structure. However, conventional self-supervised strategies are usually implemented in a Siamese-like manner. Therefore, (as a first novelty) in this study, we explore the use of Siamese-like LSSL. Another new core framework named neural ordinary differential equation (NODE). NODE is a neural network architecture that learns the dynamics of ordinary differential equations (ODE) through the use of neural networks. Many temporal systems can be described by ODE, including modeling disease progression. We believe that there is an interesting connection to make between LSSL and NODE. This paper aims at providing a better understanding of those core algorithms for learning the disease progression with the mentioned change. In our different experiments, we employ a longitudinal dataset, named OPHDIAT, targeting diabetic retinopathy (DR) follow-up. Our results demonstrate the application of LSSL without including a reconstruction term, as well as the potential of incorporating NODE in conjunction with LSSL. 

\end{abstract}

\section{Introduction}

In recent years, the deep learning community has enthusiastically embraced the self-supervised learning paradigm, by taking advantage of pretext tasks to learn better representations to be used on a downstream task. Most of the existing works are based on contrastive learning \cite{SIMCLR} or hand-crafted pretext task \cite{zhang2016colorful}. When using hand-crafted pretext tasks, the model learns automatically by obtaining supervisory signals extracted directly from the nature of the data itself, without manual annotation performed by an expert. An adequate objective function teaches robust feature representations to the model, which are needed to solve downstream tasks (e.g., classification, regression). However, to design an effective pretext task, domain-specific knowledge is required.

Recently, several approaches that involve pretext tasks in a longitudinal context have appeared with the purpose of encoding disease progression. These approaches aim to learn longitudinal changes or infer disease progression trajectories at the population or patient levels \cite{Vernhet2021,ren2022local,Zhao2021,Couronné}.

Longitudinal self-supervised learning (LSSL) was initially introduced in the context of disease progression as a pretext task by \cite{Rivail2019}, which involved the introduction of a longitudinal pretext task utilizing a Siamese-like model. The model took as input a consecutive pair of images and predicted the difference in time between the two examinations. Since then, more sophisticated algorithms have been proposed, including the advanced version of LSSL proposed in \cite{Zhao2021}. The framework attempted to theorize the notion of longitudinal pretext task with the purpose of learning the disease progression. LSSL was embedded in an auto-encoder (AE), taking two consecutive longitudinal scans as inputs. The authors added to the classic reconstruction term a cosines alignment term that forces the topology of the latent space to change in the direction of longitudinal changes.

Moreover, as conducted in \cite{Rivail2019}, \cite{kim2023learning} employed a Siamese-like architecture to compare longitudinal imaging data with deep learning. The strength of this approach was to avoid any registration requirements, leverage population-level data to capture time-irreversible changes that are shared across individuals and offer the ability to visualize individual-level changes. Neural Ordinary Differential Equations (NODEs) is a new core algorithm that has a close connection to modeling time-dependant dynamics. NODE, introduced in \cite{chen2019neural}, deals with deep learning operations defined by the solution of an ODE. Whatever the involved architecture and given an input, a NODE defines its output as the numerical solution of the underlying ordinary differential equation (ODE). One advantage is that it can easily work with irregular time series \cite{Yulia}, which is an inherent aspect of the disease progression context. This is possible because the NODEs are able to deal with continuous time. Additionally, NODEs leverage the inductive bias that time-series originates from an underlying dynamical process, where the rate of change of the current state depends on the state itself. NODEs have been used to model hidden dynamics in disease progression using neural networks. Authors in \cite{qian2021integrating} have developed a Neural ODE-based model in order to learn disease progression dynamics for patients under medication for COVID-19. Thus, Lachinov et al. proposed in \cite{lachinov2022learning} a U-Net-based model coupled with a neural ODE to predict the progression of 3D volumic data in the context of geographic atrophy using retinal OCT volumes and predicting the brain ventricle change with MRI for the quantification of progression of Alzheimer's disease.
Thus, the main objectives of our work are to examine if:

\begin{enumerate}

    \item By including Neural ODEs, it becomes possible to generate a latent representation of the subsequent scan without the explicit need for feeding an image pair to the model.  Due to the inherent characteristics of NODE, there is potential to encode the latent dynamic of disease progression and longitudinal change. We believe this established a natural connection between NODE and LSSL algorithms, warranting further investigation to gain a deeper understanding of these newly introduced frameworks.

    \item Most of the current self-supervised learning frameworks are embedded in a Siamese-like paradigm,  using only an encoder and optimizing different loss functions based on various similarity criteria. While the reconstruction term offers a way to encode anatomical change, successful longitudinal pretext tasks \cite{Rivail2019,Emre_2022,kim2023learning} have only used an encoder in terms of design, which justify the potential of extending it to LSSL. In order to examine this hypothesis, we are constructing a Siamese-like variant of the LSSL framework to evaluate the significance of the reconstruction term within LSSL.

\end{enumerate}

\section{Method}

In this section, we briefly introduce the concepts related to LSSL \cite{Zhao2021} and NODE \cite{chen2019neural}, and we investigate the longitudinal self-supervised learning framework under different scenarios: standard LSSL (Fig.\ref{fig:proposed_method}b), Siamese-like LSSL (Fig.\ref{fig:proposed_method}a) as well as their NODE-based versions (Fig.\ref{fig:proposed_method}c-d). Let $\mathcal{X}$ be the set of subject-specific image pairs extracted from the full collection of color fundus photographs (CFP). $\mathcal{X}$ contains all $(x^{t_{i}}, x^{t_{i+1}})$ image pairs that are from the same subject with image $x^{t_{i}}$ scanned before image $x^{t_{i+1}}$. These image pairs are then provided as inputs of an auto-encoder (AE) network (Fig.\ref{fig:proposed_method}). The latent representations generated by the encoder are denoted by $z^{t_i}=f(x^{t_{i}})$ and $ z^{t_{i+1}}=f(x^{t_{i+1}})$ where $f$ is the encoder. From this encoder, we can define the trajectory vector $\Delta z = (z^{t_{i+1}} - z^{t_{i}})$. The decoder $g$ uses the latent representation to reconstruct the input images such that $\tilde{x}^{t_{i}}=g(z^{t_i})$ and $\tilde{x}^{t_{i+1}}=g(z^{t_{i+1}})$. 

\subsection{Longitudinal self-supervised learning (LSSL)}

Longitudinal self-supervised learning (LSSL) exploits a standard AE (Fig.\ref{fig:proposed_method}b). The AE is trained with a loss that forces the trajectory vector $\Delta z$ to be aligned with a direction that could rely in the latent space of the AE called $\tau$. This direction is learned through a subnetwork composed of single dense layers which map dummy data (vector full of ones) into a vector $\tau$ that has the dimension of the latent space of the AE. Enforcing the AE to respect this constraint is equivalent to encouraging $\cos\left(\Delta z,\boldsymbol{\tau}\right)$ to be close to 1, i.e., a zero-angle between $\boldsymbol{\tau}$ and the direction of progression in the representation space. With $\mathbf{E}$ being the mathematical expectation, the objective function is defined as follows:

\begin{equation}
\mathbf{E}_{(x^{t_{i}}, x^{t_{i+1}}) \sim \mathcal{X}} \left(\lambda_{recon}\cdot(\parallel x^{t_{i}} - \tilde{x}^{t_{i}} \parallel_2^2 + \parallel x^{t_{i+1}} - \tilde{x}^{t_{i+1}} \parallel_2^2)-\lambda_{dir} \cdot \cos(\Delta z,\tau)\right)
\label{eq:loss_lssl}
\end{equation}

When $\lambda_{dir}=0$ and $\lambda_{recon}>0$, the architecture is reduced to a simple AE. Conversely, using $\lambda_{dir}>0$ and $\lambda_{recon}=0$ amounts to a Siamese-like structure with a cosine term as a loss function (Fig.\ref{fig:proposed_method}c).

\subsection{Neural ordinary differential equations (NODE)}

NODEs approximate unknown ordinary differential equations by a neural network \cite{NeuralODE} that parameterizes the continuous dynamics of hidden units $\mathbf{z}\in \mathbb{R}^n$ over time with $\mathbf{t}\in \mathbb{R}$. NODEs are able to model the instantaneous rate of change of $\mathbf{z}$ with respect to $\mathbf{t}$ using a neural network $u$ with parameters $\theta$.

\begin{equation} 
\label{equ:neural_odes}
\lim_{h\rightarrow0}\frac{\mathbf{z}_{t+h}-\mathbf{z}_t}{h}=\frac{d\mathbf{z}}{dh}=u(t,\mathbf{z},\boldsymbol{\theta})
\end{equation} \vspace{-0.1cm} \\

\noindent The analytical solution of Eq.\ref{equ:neural_odes} is given by:
 
\begin{equation} 
\mathbf{z}_{t_{1}} = \mathbf{z}_{t_{0}} + \int_{t_{0}}^{t_{1}}f(t,\mathbf{z},\boldsymbol{\theta})\mathrm{d}t =\textrm{ODESolve}(\mathbf{z}(t_0), u, t_0, t_1, \theta)
\label{equ:solveode}
\end{equation}

\noindent where $[t_{0}, t_{1}]$ represents the time horizon for solving the ODE, $u$ being a neural network, and $\theta$ is the trainable parameters of $u$.

\subsection{LSSL-NODE}

By using a black box ODE solver introduced in \cite{NeuralODE}, we are able to approximately solve the initial value problem (IVP) and calculate the hidden state at any desired time using Eq.\ref{equ:solveode}. We can differentiate the solutions of the ODE solver with respect to the parameters $\theta$, the initial state $\mathbf{z}_{t_0}$ at initial time $t_0$, and the solution at time $t$. This can be achieved by using the adjoint sensitivity method \cite{chen2019neural}. This allows to backpropagate through the ODE solver and train any learnable parameters with any optimizer. Typically, NODE is modelized by a feedforward layer, where we solve the ODE from $t_0$ to a terminal time $t_1$, or it can be used to output a series, by calculating the hidden state at specific times $\{t_1,...,t_i, t_{i+1}\}$. For a given patient, instead of giving a pair of consecutive images to the model, we only provide the first image of the consecutive pair. In our case, through the latent representation of this image, we define an IVP problem that aims to solve: $\dot z(t) = u(z(t), t,  \theta),$ with the initial value $z(t_{i}) = z^{{t_{i}}}$. This results in the following update of the equations from the previous notation. The latent representations generated by the encoder are denoted by $z^{t_{i}}=f(x^{t_{i}})$ and $z^{t_{i+1}}_{node}=\textrm{ODESolve}(z^
{{t_i}}, u, t_i, t_{i+1}, \theta)$ where $f$ is the encoder and $u$ is the defined neural network for our NODE. From this encoder and the NODE, we can define the trajectory vector $(\Delta_z^{node}) = (z^{t_{i+1}}_{node} - z^{t_{i}})$. The same decoder $g$ uses the latent representation to reconstruct the input images such that $\tilde{x}^{t_{i}}
=g(z^{t_{i}})$ and $\tilde{x}^{t_{i+1}}=g(z^{{t_{i+1}}}_{node})$. The objective function is defined as:

\begin{equation}
\mathbf{E}_{(x^{t_{i}}, x^{t_{i+1}}) \sim \mathcal{X}} \left(\lambda_{recon}\cdot(\parallel x^{t_{i}} - \tilde{x}^{t_{i}} \parallel_2^2 + \parallel x^{t_{i+1}} - \tilde{x}^{t_{i+1}} \parallel_2^2)-\lambda_{dir} \cdot \cos(\Delta_z^{node},\tau)\right)
\label{eq:loss_1}
\end{equation}

\begin{figure}[h]
\includegraphics[width=\textwidth]{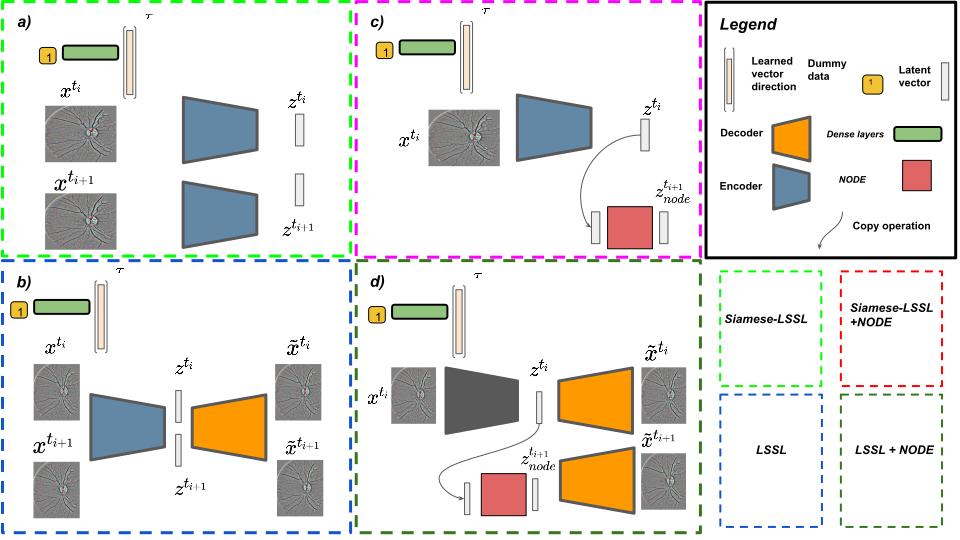}
\caption{Illustration of original LSSL and the proposed extension of the LSSL, in figure a) the Siamese-like LSSL(S-LSSL), b) original LSSL  and in figure c) and d) their NODE-based version respectively.}
\label{fig:proposed_method}
\end{figure}

\section{Experiments and Results}

\subsection{Dataset}

The proposed models were trained and evaluated on OPHDIAT \cite{ophdiat}, a large CFP database collected from the Ophthalmology Diabetes Telemedicine network consisting of examinations acquired from 101,383 patients between 2004 and 2017. Within 763,848 interpreted CFP images, 673,017 are assigned with a DR severity grade, the others being nongradable. The image sizes range from 1440 $\times$ 960 to 3504 $\times$ 2336 pixels. Each examination has at least two images for each eye. Each subject had 2 to 16 scans, with an average of 2.99 scans spanning an average time interval of 2.23 years. The age range of patients is from 9 to 91. The dataset is labeled according to the international clinical DR severity scale (ICDR) where the classes include: no apparent DR, mild non-proliferative diabetic retinopathy (NPDR), moderate NPDR, severe NPDR, and proliferative diabetic retinopathy (PDR), respectively labeled as grades 0, 1, 2, 3, and 4. NPDR (grades 1, 2, 3) corresponds to the early-to-middle stage of DR and deals with a progressive microvascular disease characterized by small vessel damages.

\noindent\textbf{Image selection.} The majority of patients from the OPHDIAT database have multiple images with different fields of view for both eyes. To facilitate the selection, we chose to randomly take a single image per eye for each examination. In addition, we defined two sub-datasets from patients that have a longitudinal follow-up to fit our experiment setup. 

\begin{enumerate}
    \item \textbf{\textit{Pair-wise dataset}}: From the OPHDIAT database, we selected pairs from patients with at least one DR severity change in their follow-up. This resulted in 10412 patients and 49579 numbers of pairs.
    \item  {\textit{\textbf{Sequence-wise dataset}}}: From the OPHDIAT database, we selected patients that have at least four visits. This resulted in 7244 patients and 13997 sequences.
\end{enumerate}

For both datasets, the split was conducted with the following distribution: training (60\%), validation (20\%), and test (20\%) based on subjects, i.e., images of a single subject belonged to the same split and in a way that preserves the same proportions of examples in each class as observed in the original dataset.  We also ensured that for both datasets, there was no intersection between a patient in training/validation/test sets. Except for the registration stage, we followed the same image processing performed in \cite{zeghlache}.

% \noindent\textbf{Image pre-processing.} Before the training the image were adaptively cropped to the width of the field of view (i.e., the eye area in the CPF image) and then resized to 256$\times$256. A Gaussian filter estimates the background in each color channel to attenuate the strong intensity variations among the dataset which is then subtracted from the image. Finally, the field of view is eroded by 5\% to eliminate illumination artifacts around the edges. Due to the design image selection no image registration was performed between consecutive pairs. 

\subsection{Implementation details} 

As conducted in \cite{Zhao2021,Ouyang,zeghlache}, a basic AE was implemented in order to focus on the methodology presented. We used the same encoder and decoder as used in \cite{zeghlache}. This encoder provides a latent representation of size 64 × 4 × 4. The different networks were trained for 400 epochs by the AdamW optimizer, with a learning rate of $5 \times 10^{-4}$, OneCycleLR as scheduler, and a weight decay of $10^{-5}$, using an A6000 GPU with the PyTorch framework and trained on the pair-wise dataset. The regularization weights were set to $\lambda_{dir}=1.0$ and $\lambda_{recon}=1.0$. Concerning NODE, we used the torchdiffeq framework, which provides integration of NODE in Pytorch with the possibility to use a numeric solver with the adjoint method, which allows constant memory usage. Our neural ODE function consists of the succession of dense layers followed by a tanh activation function. We employed the fifth-order "dopri5" solver with an adaptive step size for the generative model, setting the relative tolerance to 1e-3 and the absolute tolerance to 1e-4. When using the dopri solver with the adjoint method, the torchdiffeq package does not provide support for a batched time solution, so during training for the forward batch in the NODE we used the workaround introduced in \cite{lechner2020longterm} to enable a batched time solution. 

\subsection{Evaluating the learned feature extractor}

To evaluate the quality of the encoder from the different experiments, we will use two tasks with a longitudinal nature. We only perform fine-tuning scenarios, as performed in \cite{Emre_2022} to assess the quality of the learned weights. We initialize the weights of the encoder that will be used for the two tasks, with the weights of the longitudinal pre-training method presented in Fig.\ref{fig:proposed_method}. An additional AE and AE-NODE were added for comparison purposes. Each task was trained for 150 epochs with a learning rate of $10^{-3}$ and a weight decay of $10^{-4}$ and OneCycleLR as a scheduler. The first one is the prediction of the patient's age in years, using a CFP (this task is called \textit{\textbf{age regression}} for the rest of the manuscript), and the second task is the prediction of the development of the DR for the next visit using the past three examinations (task called \textbf{\textit{predict next visit}}). For the age regression, the model developed is a combination of the implemented encoder and a multi-layer perceptron. The model is trained with the Mean Squared Error (MSE) on the image selected for the sequence-wise dataset. For predict next visit, we used a CNN+RNN \cite{CUI20191}, a standard approach for tackling sequential or longitudinal tasks. The RNN we used is a long short-term memory (LSTM). The model is trained using cross-entropy, on the sequence-wise dataset. We used the Area Under the receiver operating characteristic Curve (AUC) with three binary tasks which are the following: predicting at least NPDR in the next visit (\textbf{AUC Mild+}), at least moderate NPDR in the next visit denoted (\textbf{AUC Moderate+}) and finally at least severe NPDR in the next visit (\textbf{AUC severe+}) for evaluating the models.

\begin{table}[h!]
\begin{adjustbox}{width=5.5cm,center}

\begin{tabular}{|l|c|c|c|c|}
\hline
Weights & $\lambda_{dir}$ & $\lambda_{recon}$ & NODE & MSE \\ \hline
From scratch & -   & -   & -    &     0.0070  \\ \hline
  AE       & 0    & 1    & No   &    0.0069  \\ \hline
  AE       & 0    & 1    & Yes  &    0.0069   \\ \hline
  LSSL     & 1    & 1    & No   &     \underline{0.0050}  \\ \hline
  LSSL     & 1    & 1    & Yes  &      \textbf{0.0048}  \\ \hline
  Siamese LSSL    & 1    & 0    & No  & \underline{0.0049}  \\ \hline
  Siamese LSSL    & 1    & 0    & Yes  & \textbf{0.0046}\\ \hline
\end{tabular}
\end{adjustbox}
\caption{Results for the age regression task reporting the Mean Squared Error (MSE) in squared years. Best results for the LSSL and S-LSSL are in underline, best results for their NODE version are in bold.}
\label{table:1}
\end{table}

% 0.0070136995054781
% 0.0069468962028622
% 0.0069240457378327

% 0.0050544892437756
% 0.0048652179539203

% 0.0049592009745538
% 0.0046595460735261

\begin{table}[h!]
\begin{adjustbox}{width=9cm,center}\begin{tabular}{|l|l|l|l|l|l|l|l|}
\hline
Weights & $\lambda_{dir}$ & $\lambda_{recon}$ & NODE &  AUC Mild+ & AUC Moderate + & AUC severe+ \\ \hline
From scratch & -   & -   & -   & 0.574 & 0.602& 0.646  \\ \hline
  AE       & 0    & 1    & No   &0.563 & 0.543& 0.636   \\ \hline
  AE       & 0    & 1    & Yes & 0.569& 0.565& 0.649  \\ \hline
  LSSL     & 1    & 1    & No   & \underline{0.578} & \underline{0.618}& \underline{0.736} \\ \hline 
  LSSL     & 1    & 1    & Yes & \textbf{0.604} & \textbf{0.630}&\textbf{0.760}   \\ \hline
  Siamese LSSL    & 1    & 0    & No & \underline{0.578} & \underline{0.596} & \underline{0.708}   \\ \hline
  Siamese LSSL    & 1    & 0    & Yes & \textbf{0.569} & \textbf{0.549} & \textbf{0.756}\\ \hline
\end{tabular}
\end{adjustbox}
\caption{Results for the predict next visit label for the different longitudinal pretext task for the three binary tasks using the AUC. Best results for the LSSL and S-LSSL are in underline, best results for their NODE version are in bold.}
\label{table:2}

\end{table}

According to the results of both Tab.\ref{table:1} and \ref{table:2}, we observe that longitudinal pre-training is an efficient pre-training strategy to tackle a problem with a longitudinal nature compared to training from scratch of classic autoencoder, which is aligned with the following studies \cite{Emre_2022,Ouyang,Zhao2021,zeghlache}. For the age regression task, according to the results presented in Tab.\ref{table:1}, the best longitudinal pre-training strategy is the Siamese LSSL version with NODE. However, the difference in performance is marginal, indicating that the Siamese LSSL could also be used as pretext task. The cosine alignment term in the loss function, which is the one responsible for forcing the model to encode longitudinal information, seems to be beneficial in solving longitudinal downstream tasks. 

For the predict next visit task (Tab.\ref{table:2}), the LSSL-NODE version performs better than the rest. An observation that could explain the difference is that during the training of the LSSL-NODE, the reconstruction term and the direction alignment (second term of Eq.\ref{eq:loss_1}) in the loss function reache 0 at the end of the training. While for the LSSL, only the reconstruction term converges to 0. The direction alignment term faces a plateau of around 0.3. One way to explain this convergence issue is the fact that the longitudinal pair were selected randomly. Even if it is the same patient, the image may have a very different field of view which could increase the difficulty for the model to both minimize the reconstruction loss and cosine alignment term. The LSSL-NODE does not have this issue since only one image is given to the CNN, which we suspect ease the problem solved by the LSSL+NODE. Another way to explain this phenomena is the fact that the LSSL was trained with $\lambda_{recon}$ and $\lambda_{dir}$ set to 1, a more advanced weight balanced between the two terms in the loss could reduce this issue. This observation could be the explanation for the difference in performance when using the LSSL-NODE vs classic LSSL. Another simple explanation could be the fact that we did not use any registration method for the longitudinal pairs. Surprisingly, according to both Tab.\ref{table:1} and \ref{table:2}, the model with the NODE extension performed well compared to their original version. We believe that the NODE forces the CNN backbone to learn a representation that is more suitable for modeling the dynamics of the disease progression.

% 0.5638242894056847,0.5432310662947866,0.6364946292900183 AE 
% 0.5695325346488137,0.5654795108345849,0.6496812505458038 AE NODE

% 0.5782632526818573,0.5966487878137738,0.7080604314033709 Siamese-LSSL 
% 0.5691436327095242,0.5499678180647929,0.7567461357086718, Siamese-LSSL + NODE

%0.578888366872863,0.6182879210469857,0.7366168893546415 LSSL
%0.6043510035758097,0.6305942930701566,0.7600646231770151 LSSL NODE

% 0.6467120775478123 from scratch

%0.7366168893546415 LSSL
% 0.6909003580473321
% 0.7042179722294997

% Only the method with the cos alignement term outperform the baseline. The AE

\subsection{Analysis of the norm of $\Delta_{z}$}

Using the same protocol introduced in \cite{Ouyang,Zhao2021}, we computed the norm of the trajectory vector. The intuition behind this computation is the following: $\Delta_z$ can be seen as some kind of vector that indicates the speed of disease progression because it can be regarded as an instantaneous rate of change of $\mathbf{z}$ normalized by 1. For the different extensions of LSSL, we evaluate pregnancy factor \cite{diabetexpregnancy} and type of diabetes \cite{Chamard2020}, which are known factors that characterize the speed of the disease progression in the context of DR. This is done to analyze the capacity of $\Delta_z$ to capture the speed of disease progression.

\begin{enumerate}
    \item \textbf{\textit{Pregnancy}}: Pregnant vs not pregnant (only female). For the pregnant group, we selected longitudinal pairs from patients that were in early pregnancy and in close monitoring. For the rest, we selected longitudinal pair from female patients without antecedent of pregnancy.  
     \item \textit{\textbf{Diabetes type}:} Patients with known diabetes type. We only selected a longitudinal pair of patients for known diabetes type 1 or 2. 

\end{enumerate}

First, patients present in the training set for the longitudinal pretext task were not allowed to be selected as part of any group. For the first group, we randomly selected 300 patients for each category. For the second group, we randomly selected 2000 patients for the two categories. We applied a statistical test (student t-test) to explore if the mean value of the norm of the trajectory vector with respect to both defined factors has a larger mean value for patients with pregnancy than for patients without pregnancy. And if the patient with type 1 diabetes had a higher mean than type 2. We observe that the norm of the trajectory vector $(\Delta z)$ is capable of dissociating the two types of diabetes (t-test p-value < 0.01) and pregnancy type (t-test p-value < 0.00001) for all models that have a cosine alignment term (except the AE and the AE+NODE extension). Regarding the type of diabetes, a specific study in the OPHDIAT dataset \cite{Chamard2020} indicated that the progression of DR was faster in patients with type 1 diabetes than for patients with type 2 diabetes. Furthermore, pregnancy is a known factor \cite{diabetexpregnancy} in the progression speed of DR. Those observations are aligned with the expected behavior of the two factors. In addition, as observed in \cite{Ouyang,zeghlache,Zhao2021}, standard AE is not capable of encoding disease progression.

\subsection{Evaluation of the NODE weights}

\begin{figure}[h!]
\includegraphics[width=10cm]{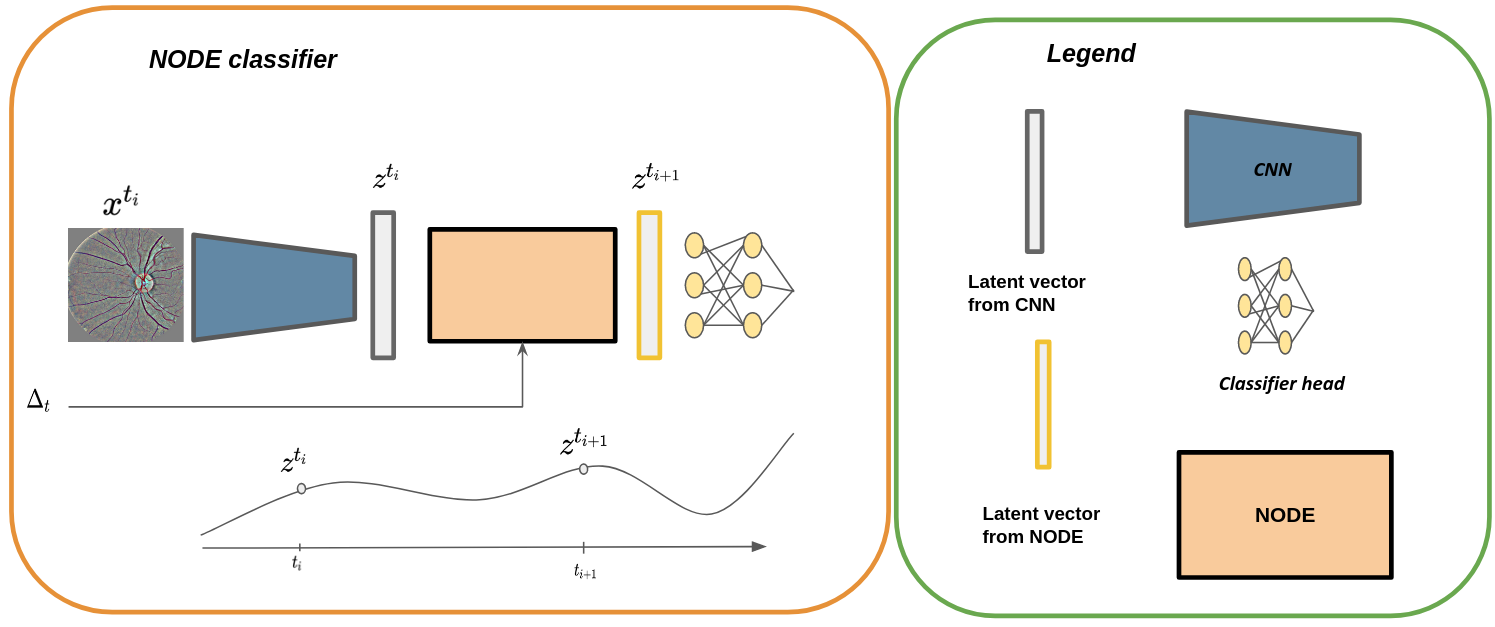}
\caption{Illustration of proposed NODE classifier for evaluating the weights of the method with a backbone + NODE}
\label{fig:node_classifier}
\end{figure}

For models trained with a NODE, we design a protocol to assess the capacity of NODE to learn a meaningful representation related to disease progression. The protocol to evaluate the weights of the NODE is as follows. We implemented a NODE classifier (NODE-CLS) illustrated in Fig.\ref{fig:node_classifier}. The NODE classifier is constructed as the concatenation of a backbone, a NODE, and a multilayer perceptron (MLP). The architecture of the backbone and the NODE is the same that was used with the LSSL method. The MLP consists of two fully connected layers of dimensions 1024 and 64 with LeakyReLU activation followed by a last single perceptron, which project the last layer representation into the number of class, and the network is trained using the pairwise dataset. The NODE-CLS only requires a single CFP denoted $x^{t_{i}}$ and $\Delta_{t} = t_{i+1} - t_{i}$, which is the time lapse between examinations $x^{t_{i}}$ and $x^{t_{i+1}}$. The image is first given to the backbone, which produces the latent representation $z^{t_{i}}$, then this vector is fed to the black box NODE, using the same IPV defined for the LSSL-NODE we can define the latent representation of the next visit. Using the predicted latent representation by the NODE, we predicted the severity grade of the next visit with the MLP. The same loss and metrics were used for the predict next visit task.

% Like for the predict next visit task the model is trained with the cross entropy and evaluated with the AUC for the three binary tasks. 

\begin{table}[]
\centering
\begin{adjustbox}{width=9cm,center}\begin{tabular}{|l|l|l|l|l|l|l|}
\hline
Weights & $\lambda_{dir}$ & $\lambda_{recon}$ & NODE & AUC Mild+ & AUC moderate+ & AUC severe+  \\ \hline
From scratch & -   & -   & -    &  0.552 & 0.600 & 0.583  \\ \hline
  AE     & 0    & 1    & Yes  &   0.561 & 0.609 & 0.590 \\ \hline
  LSSL     & 1    & 1    & Yes  & \textbf{0.547}  & \textbf{0.609} & \textbf{0.641}  \\ \hline
  Siamese LSSL    & 1    & 0    & Yes &\textbf{0.558}& \textbf{0.617} & \textbf{0.670 } \\ \hline
\end{tabular}
\end{adjustbox}
\caption{Comparaison of AUC for the NODE classifier using different initialize weights, best results in bold.}
\label{table:node_classifier}
\end{table}

%0.5529442233377833,0.6000436444300636,0.583076245493281,
%0.5615733401958805,0.6093691509863743,0.590688503769256,

%0.5471490346266733,0.6098484104898262,0.6414351851851852,
%0.5581432580538126,0.6175119933336898,0.6704389954113404
%
%
% ,0.5200624509658499,0.5616851996136139,0.5513325549000327,NODE_classifier_version_5305

% ,0.6098484104898262,0.6414351851851852,NODE_classifier_version_15564_from_S_LSSL_version_5406

% 0.6175119933336898,0.6704389954113404,NODE_classifier_version_5323

For the NODE-CLS, Tab.\ref{table:node_classifier} shows a clear performance gain compared to the method trained from scratch. The best weights are obtained with the Siamese-LSSL. The classical LSSL also provides descent result, for the different subtask. We suspect that the LSSL model is more affected by the design of our experiments. Since we randomly selected images per examination, we did not perform any registration step. These results suggest that the classical LSSL is more likely to underperform than the Siamese-like method when the pair are not registered, aligned with the finding of \cite{kim2023learning}.

\section{Discussion and conclusion}

In this paper, we investigated the use of the LSSL under different scenarios in order to get a better understanding of the LSSL framework. Our result demonstrated that the use of Siamese-like LSSL is possible. The S-LSSL might be more suitable when not registered pairs are given to the model. Moreover, for the various tasks that were used to evaluate the quality of the weights, a performance gain was observed when the LSSL and S-LSSL coupled with the NODE compared to their standard versions, while the reasons for such performance remain unclear. We hypothesize that adding the NODE in training forces the linked backbone to provide an enriched representation embedded with longitudinal information.  Thus, since the formulation of the original LSSL \cite{Zhao2021} is based on a differential equation, which resulted in the introduction of the cosine alignment term in the objective function. Including the NODE is aligned with the formulation of the LSSL problem, which is reflected in our results. In this direction,  we are interested in looking at the representation of the learned neural ODE, in order to understand if the neural ODE is able to interpolate the spatial feature and thus give the opportunity not even to need the registration. If so, the use of NODE with longitudinal self-supervised learning could also overcome the need of heavy registration step. LSSL techniques are quite promising and we believe that they will continue to grow at a fast pace. We would like to extend this study by including more frameworks \cite{Emre_2022,kim2023learning,Ouyang,Couronné} as well as more longitudinal datasets.

Some limitations should be pointed out, since we selected a random view of CFP for each examination, we did not apply any registration step. Using a better pairing strategy with the right registration step could enhance the results presented for LSSL. Moreover, we did not perform any elaborate hyperparameter search to find the correct loss weights in order to reach a complete loss convergence for the LSSL.

This work also opens up interesting research questions related to the use of neural ODE. The formulation and hypotheses of the LSSL framework are based on a differential equation that could be directly learned via a Neural ODE which could explain the better performance of the LSSL+NODE version. In the future, we will work on a theoretical explanation as to why this blending works.

In addition, for the models trained with a NODE, our experiments demonstrated the possibility of using pretext tasks on Neural ODE in order to provide an enhanced representation. In the classic SSL pretext task paradigm, the goal is to learn a strong backbone that yields good representation to solve specific downstream task. The results from the NODE-CLS experiments indicated that pre-training techniques where a NODE is involved can be reused on longitudinal downstream tasks and have the ability to enhance the results when a NODE is part of the model. In the future, we would like to explore time-aware pretraining in general, with a pretext task that has longitudinal context to be reused on longitudinal tasks.  \\

\noindent\textbf{Acknowledgements}
The work takes place in the framework of Evired, an ANR RHU project. This work benefits from State aid managed by the French National Research Agency under the ``Investissement d'Avenir'' program bearing the reference ANR-18-RHUS-0008.

\bibliographystyle{splncs04}
\bibliography{biblio}

\end{document}